\def\year{2020}\relax
\documentclass[letterpaper]{article} 
\usepackage{aaai20}  
\usepackage{times}  
\usepackage{helvet} 
\usepackage{courier}  
\usepackage[hyphens]{url}  
\usepackage{graphicx} 
\usepackage{graphicx}  
\usepackage{graphicx}
\usepackage{subfigure}
\usepackage{amssymb}
\usepackage{amsmath}
\usepackage{booktabs}       
\usepackage{amsfonts}       
\usepackage{multirow}
\usepackage{lipsum}
\usepackage{array}
\usepackage{pgffor}
\newcolumntype{C}[1]{>{\centering\arraybackslash}m{#1}}
\title{Natural Image Matting via Guided Contextual Attention}
\author{Yaoyi Li, Hongtao Lu\thanks{Corresponding author} \\
Department of Computer Science and Engineering, Shanghai Jiao Tong University, China\\
dsamuel@sjtu.edu.cn, htlu@sjtu.edu.cn
}
 
\begin{document}

\maketitle

\begin{abstract}
Over the last few years, deep learning based approaches have achieved outstanding improvements in natural image matting. Many of these methods can generate visually plausible alpha estimations, but typically yield blurry structures or textures in the semitransparent area. This is due to the local ambiguity of transparent objects. One possible solution is to leverage the far-surrounding information to estimate the local opacity. Traditional affinity-based methods often suffer from the high computational complexity, which are not suitable for high resolution alpha estimation. Inspired by affinity-based method and the successes of contextual attention in inpainting, we develop a novel end-to-end approach for natural image matting with a guided contextual attention module, which is specifically designed for image matting. Guided contextual attention module directly propagates high-level opacity information globally based on the learned low-level affinity. The proposed method can mimic information flow of affinity-based methods and utilize rich features learned by deep neural networks simultaneously. Experiment results on Composition-1k testing set and alphamatting.com benchmark dataset demonstrate that our method outperforms state-of-the-art approaches in natural image matting. Code and models are available at https://github.com/Yaoyi-Li/GCA-Matting.
\end{abstract}

\section{Introduction}
The natural image matting is one of the important tasks in computer vision. It has a variety of  applications in image or video editing, compositing and film post-production \cite{wang2008image,aksoy2017designing,lutz2018alphagan,xu2017deep,samplenet}.
 Matting has received significant interest from the research community and been extensively studied in the past decade. 
 Alpha matting refers to the problem that separating a foreground object from the background and estimating transitions between them. The result of image matting is a prediction of alpha matte which represents the opacity of a foreground at each pixel.

\begin{figure}[t]
	\centering
	\includegraphics[width = 0.95\columnwidth]{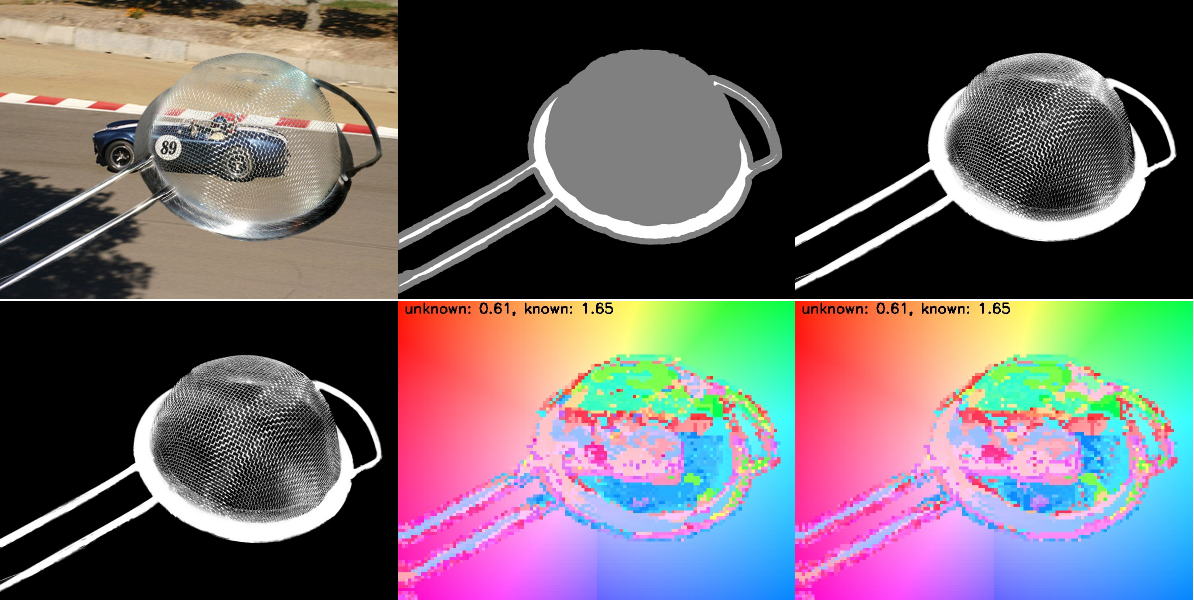}
	\caption{The visualization of our guided contextual attention map. Top row from left to right, the image, trimap and ground-truth. Second row, the alpha matte prediction, attention offset map from first GCA block in the encoder, offset from GCA block in the decoder.}
	\label{fig:offset}
\end{figure}

Mathematically, the natural image $ I $ is defined as a convex combination of foreground image $ F $ and  background image $ B $ at each pixel $ i $ as:
\begin{equation}
	I_i = \alpha_iF_i + (1-\alpha_i)B_i, \quad \alpha_i \in [0,1],
	\label{eq:matting}
\end{equation}
where $ \alpha_i $ is the alpha value at pixel $ i $ that denotes the opacity of the foreground object. If $ \alpha_i $ is not $ 0 $ or $ 1 $, then the image at pixel $ i $ is mixed. Since the foreground color $ F_i $, background color $ B_i $ and the alpha value  $ \alpha_i $ are left unknown, the expression of alpha matting is ill-defined. Thus, most of the previous hand-crafted algorithms impose a strong inductive bias to the matting problem. 

One of the basic idea widely adopted in both affinity-based and sampling-based algorithms is to borrow information from the image patches with similar appearance. Affinity-based methods \cite{levin2008closed,chen2013knn,aksoy2017designing} borrow the opacity information from known patches with the similar appearance to unknown ones. Sampling-based approaches \cite{wang2007optimized,gastal2010shared,he2011global,feng2016cluster} borrow a pair of samples from the foreground and background to estimate the alpha value at each pixel in the unknown region based on some specific assumption.
One obstacle of the previous affinity-based and sampling-based methods is that they cannot handle the situation that there are only background and unknown areas in the trimap. It is because that these methods have to make use of both foreground and background  information  to estimate the alpha matte.

Benefiting from the Adobe Image Matting dataset \cite{xu2017deep}, more learning-based image matting methods \cite{xu2017deep,lutz2018alphagan,lu2019indices,samplenet} has emerged in recent years.
Most of learning-based approaches use network prior as the inductive bias and predict alpha mattes directly. Moreover, SampleNet \cite{samplenet} proposed to leverage deep inpainting methods to generate foreground and background pixels in the unknown region rather than select from the image. It provides a combination of the learning-based and sampling-based approach. 

In this paper, we propose a novel image matting method based on the opacity propagation in a neural network.
The information propagation has been widely adopted within the neural network framework in recent years, from natural language processing \cite{vaswani2017attention,yang2019xlnet}, data mining \cite{kipf2016semi,velivckovic2017graph} to computer vision \cite{yu2018generative,wang2018non}. 
SampleNet Matting \cite{samplenet} indirectly leveraged the contextual information for foreground and background inpainting. In contrast, our proposed method conducts information flow from the image context to unknown pixels directly. We devise a guided contextual attention module, which mimic the affinity-based propagation in a fully convolutional network. In this module, the low-level image features are used as a guidance and we perform the alpha feature transmission based on the guidance. We show an example of our guided contextual attention map in Figure \ref{fig:offset} and more details in the section of results. In the guided contextual attention module, features from two 
distinct
network branches are leveraged together. The information of both known and unknown patches are transmitted to 
feature patches in the unknown region with similar appearance.

\begin{figure*}[t]
	\centering
	\includegraphics[width = 0.9\textwidth]{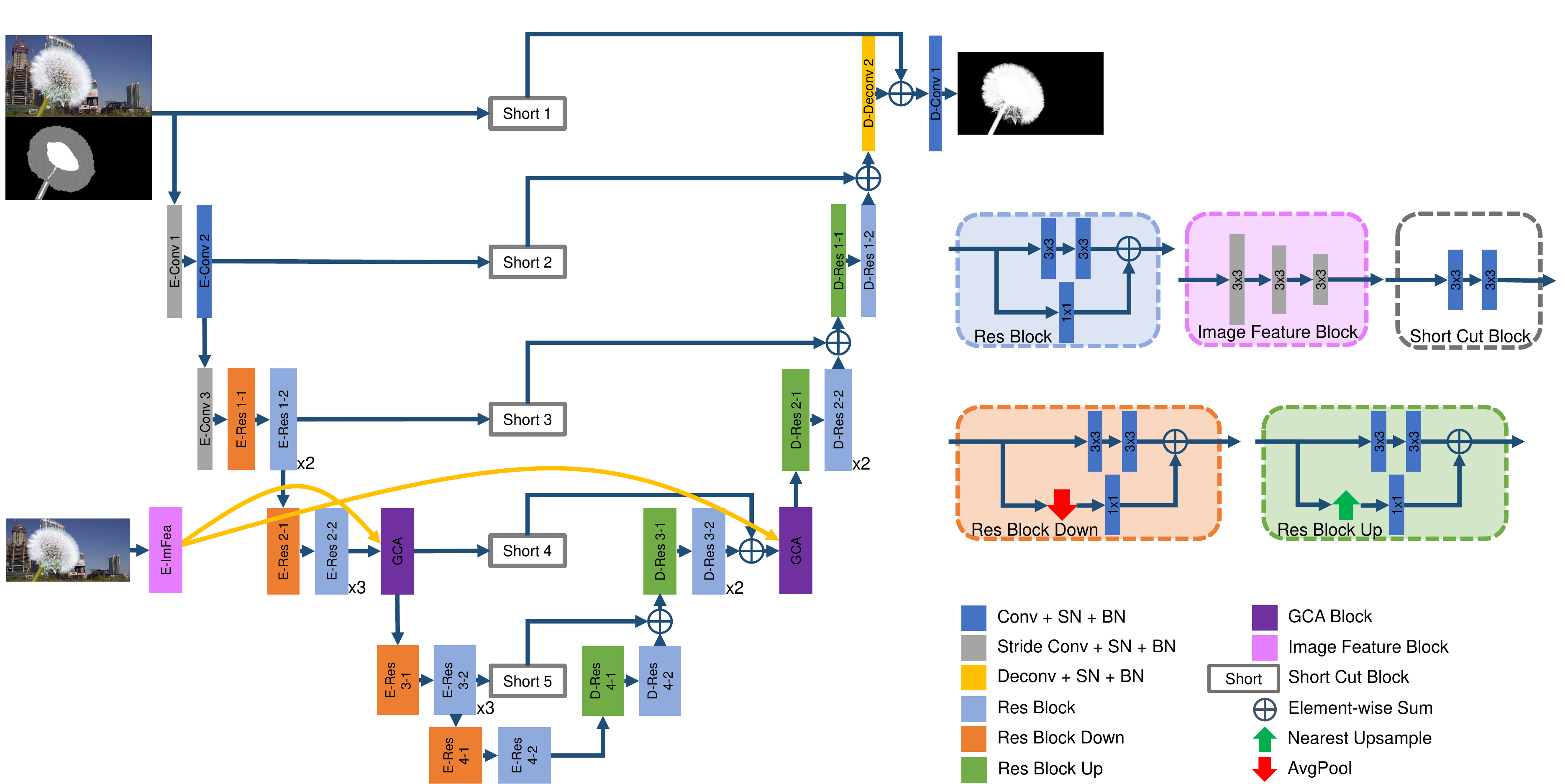}
	\caption{Overview of our proposed guided contextual attention matting framework. The baseline model shares the same architecture without GCA blocks and image feature block. Original image and trimap are the inputs of alpha feature. Image feature block and GCA blocks only takes the original merged image as input. The blue arrows denote alpha feature flow and yellow arrows denote low-level image feature flow. GCA: guided contextual attention; SN: spectral normalization; BN: batch normalization; $ \times N $: replicate $ N $ times.}
	\label{fig:arch}
\end{figure*}

Our proposed method can be viewed from two different perspectives. On one hand, the guided contextual attention can be elucidated as an affinity-based method for alpha matte value transmission with a network prior. Unknown patches share high-level alpha features with each other under the guidance of similarity between low-level image features.
On the other hand, the proposed approach can also be seen as a guided inpainting task. In this aspect, image matting task is treated as an inpainting task on the alpha image under the guidance of 
input  image. The unknown region is analogous to the holes to be filled in image inpainting. Unlike inpainting methods which borrows pixels from background of the same image, image matting borrows pixel value $ 0 $ or $ 1 $ from the known area in the alpha matte image under the guidance of original RGB  image to fill in the unknown region.

\section{Related Work}

In general, natural image matting methods can be classified into three categories: sampling-based methods, propagation methods and learning-based methods.

Sampling-based methods \cite{wang2007optimized,gastal2010shared,he2011global,feng2016cluster} solve combination equation \eqref{eq:matting} by sampling colors from foreground and background regions for each pixel in the unknown region. The pair of foreground and background samples are selected under different metrics and assumptions. Then the initial alpha matte value is calculated by the combination equation. Robust Matting \cite{wang2007optimized} selected   samples along the boundaries with confidence. The matting function was optimized by a Random Walk. Shared Matting \cite{gastal2010shared} selected the best pairs of samples for a set of neighbor pixels and reduced much of redundant computation cost. In Global Matting \cite{he2011global}, all samples available in image were utilized to estimate the alpha matte. The sampling was achieve by a randomized patch match algorithm.
More recently, CSC Matting \cite{feng2016cluster} collected a set of more representative samples by sparse coding to avoid missing out true sample pairs.

Propagation methods \cite{levin2008closed,chen2013knn,aksoy2017designing}, which are also known as affinity-based methods, estimate alpha mattes by propagating the alpha value from foreground and background to each pixel in the unknown area. The Closed-form Matting \cite{levin2008closed} is one of the most prevailing algorithm in propagation-based methods. It solved the cost function under the constraint of local smoothness. KNN Matting \cite{chen2013knn} collected matching nonlocal neighborhoods globally by K nearest neighbors. 
Moreover, the Information-flow Matting \cite{aksoy2017designing} proposed a color-mixture flow which combined the local and nonlocal  affinities of colors and spatial smoothness.

Due to the tremendous success of deep convolutional neural networks, learning-based methods achieve a dominate position in recent natural image matting \cite{cho2016natural,xu2017deep,lutz2018alphagan,lu2019indices,samplenet}.
DCNN Matting \cite{cho2016natural} is the first method that introduced a deep neural network into image matting task. It made use of the network to learn a combination of results from different previous methods. Deep Matting \cite{xu2017deep} proposed a fully neural network model with a large-scale dataset for learning-based matting methods, which was one of the most significant work in deep image matting. Following Deep Matting, AlphaGan \cite{lutz2018alphagan} explored the deep image matting within a generative adversarial framework. More subsequent work like SampleNet Matting \cite{samplenet} and IndexNet \cite{lu2019indices} with different architectures also yielded appealing alpha matte estimations.

\section{Baseline Network for Deep Image Matting}
Our proposed model uses the guided contextual attention module and a customized U-Net \cite{ronneberger2015u} architecture to perform deep natural image matting. We first construct our customized U-Net baseline for matting, then introduce the proposed guided contextual attention (GCA) module.

\subsection{Baseline Structure}
The U-Net \cite{ronneberger2015u} like architecture 
are prevailing in recent 
 matting tasks \cite{lutz2018alphagan,samplenet,lu2019indices} as well as image segmentation \cite{long2015fully}, image-to-image translation \cite{isola2017image} and image inpainting \cite{liu2018image}. 
Our baseline model shares almost the same network architecture with 
guided contextual attention
 framework in Figure \ref{fig:arch}. The only difference is that the baseline model replaces GCA blocks with identity layers and has no image feature block. 
The input to this baseline network is a cropped image patch and a 3-channel one-hot trimap which are concatenated as a 6-channel input. The output is corresponding 
estimated alpha matte.
The baseline structure is built as an encoder-decoder network with stacked residual blocks \cite{he2016deep}. 

Since the low-level features play a crucial role in retaining the detailed texture information in alpha mattes, in our customized baseline model, the decoder combines encoder features just before upsampling blocks instead of after each upsampling block. Such a design can avoid more convolutions on the encoder features, which are supposed to provide lower-level feature. We also use a two layer short cut block to align channels of encoder features for feature fusion. Moreover, in contrast to the typical U-Net structure which only combines different middle-level features, we directly forward the original input to the last convolutional layer through a short cut block instead. These features do not share any computation with the stem. Hence, this short cut branch only focuses on detailed textures and gradients.

In addition to the widely used batch normalization \cite{ioffe2015batch}, we introduce the spectral normalization \cite{miyato2018spectral} to each convolutional layer to add a constraint on Lipschitz constant of the network and stable the training, which is prevalent in image generation tasks \cite{brock2018large,zhang2018self}.

\subsection{Loss Function}
Our network only leverages one alpha prediction loss. The alpha prediction loss is defined as an absolute difference between predicted and ground-truth alpha matte averaged over the unknown area:
\begin{equation}
\mathcal{L} = \frac{1}{|\mathcal{U}|} \sum_{i\in \mathcal{U}}|\hat{\alpha_i}-\alpha_i|,
\end{equation}
where $ \mathcal{U} $ indicates the region labeled as unknown in the trimap, $ \hat{\alpha_i}$ and  $\alpha_i $ denote the predicted and ground-truth value of alpha matte as position $ i $.

There are some losses proposed in prior work for the deep image matting tasks, like compositional loss \cite{xu2017deep}, gradient loss \cite{samplenet} and Gabor loss \cite{li2019inductive}. 
Compositional loss used in Deep Matting \cite{xu2017deep} is the absolute difference between the original input image and predicted image composited by the ground-truth foreground, background and the predicted alpha mattes.
The gradient loss calculates the averaged absolute difference between the gradient magnitude of predicted and ground-truth alpha mattes in the unknown region.
Gabor loss proposed in \cite{li2019inductive} substitutes the gradient operator with a bundle of Gabor filters and aims to have a more comprehensive supervision on textures and gradients than gradient loss.

We delve into these losses to reveal whether involving different losses can benefit the alpha matte estimation in our baseline model. We provide an ablation study on Composition-1k testing set \cite{xu2017deep} in Table \ref{tab:ablation}. As Table \ref{tab:ablation} shows, the use of compositional loss does not bring any notable difference under MSE and Gradient error, and both errors increase when we  incorporate the gradient loss and alpha prediction loss. Although the adoption of Gabor loss can reduce the Gradient error to some degree, it also slightly increases the MSE.
Consequently, we only opt for the alpha prediction loss in our model.

\begin{table}[t]
	\caption{Ablation study on data augmentation and different loss functions with baseline structure. The quantitative results are tested on Composition-1k testing set. Aug: data augmentation; Rec: alpha prediction loss; Comp: compositional loss; GradL: gradient loss; Gabor: Gabor loss.}
	\centering
	\small
	\begin{tabular}{ccccc|cc}  
		\toprule
		Aug&Rec&Comp&GradL&Gabor& MSE & Grad\\
		\midrule
		\checkmark&\checkmark&&& & 0.0106 & 21.53 \\  
		\checkmark&\checkmark&\checkmark&&& 0.0107  & 21.85\\
		\checkmark&\checkmark&&\checkmark& & 0.0108 & 22.51\\ 
		\checkmark&\checkmark&&&\checkmark & 0.0109 & 20.66\\ 
		&\checkmark&&& & 0.0146 & 32.01 \\
		\bottomrule
	\end{tabular}
	\label{tab:ablation}
\end{table}

\subsection{Data Augmentation}
Since the most dominant image matting dataset proposed by \citeauthor{xu2017deep} only contains 431 foreground objects for training. We treat the data augmentation as a necessity of our baseline model.
We introduce a sequence of data augmentation.

Firstly, following the data augmentation in \cite{samplenet}, we randomly select two foreground object images with a probability of 0.5 and combine them to obtain a new foreground object as well as a new alpha image. Subsequently, the foreground object and alpha image will be resized to $ 640 \times640 $ images with a probability of 0.25. In this way, the network can nearly see the whole foreground image instead of a cropped snippet. Then, a random affine transformation are applied to the foreground image and the corresponding alpha image. We define a random rotation, scaling, shearing as well as the vertical and horizontal flipping in this affine transformation.  Afterwards, trimaps are generated by a dilation and an erosion on alpha images with random number of pixels ranging from 5 to 29. With the trimap obtained, we randomly crop one $ 512\times512 $ patch from each foreground image, corresponding alpha and trimap respectively. All of the cropped patches are centered on an unknown region. The foreground images are then converted to HSV space, and different jitters are imposed to the hue, saturation and value.
Finally, we randomly select one background image from MS COCO dataset \cite{lin2014microsoft} for each foreground patch and composite them to get the input image.

To demonstrate the effectiveness of data augmentation, we conduct an experiment with minimal data augmentation. In this case, only two necessary operations, image cropping and trimap dilation are retained. More augmentations like random image resize and flipping, which are widely used in most of previous deep image matting methods \cite{xu2017deep,lutz2018alphagan,samplenet,lu2019indices}, are not included in this experiment. We treat this experiment setting as no data augmentation. The experimental results are also listed in Table \ref{tab:ablation}. We can see that without additional augmentation, our baseline model already achieves comparable performance with Deep Matting.

\section{Guided Contextual Attention Module}

The guided contextual attention module contains two kinds of components, an image feature extractor block for low-level image feature and one or more guided contextual attention blocks for information propagation.

\begin{figure}[t]
	\centering
	\includegraphics[width = 0.95\columnwidth]{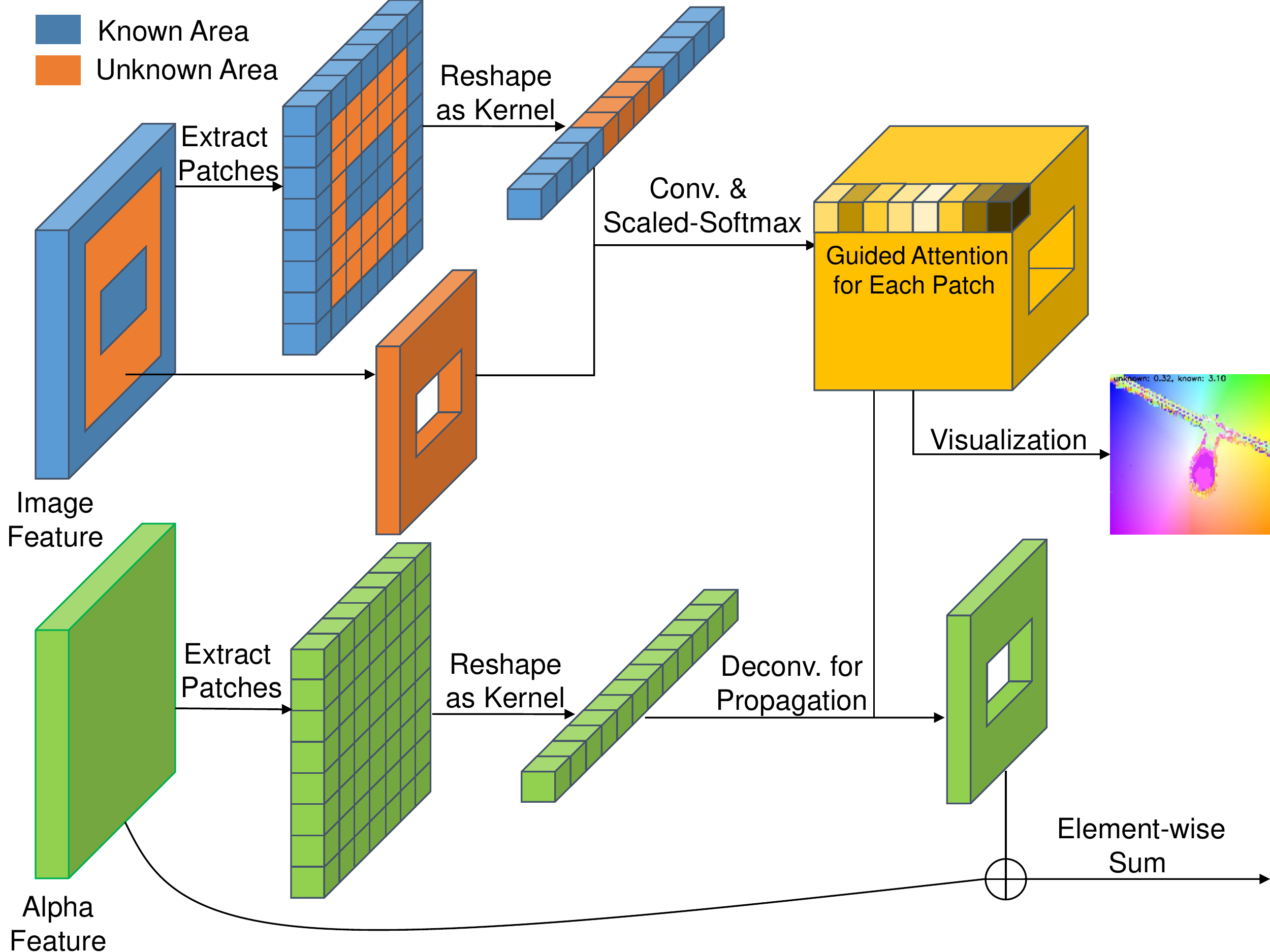}
	\caption{The illustration of the guided contextual attention block. Computation is implemented as a convolution or a deconvolution. Two additional $ 1\times1 $ convolutional layers for adaptation are not shown in this figure to keep neat. One is applied to the input image feature before extracting patches, and the other one is applied to the result of propagation before the element-wise summation.}
	\label{fig:gca}
\end{figure}

\subsection{Low-level Image Feature}

Most of the affinity-based approaches have a basic inductive bias that local regions with almost identical appearance should have similar opacity. This inductive bias allows the alpha value propagates from the known region of a trimap to the unknown region based on affinity graph, which can often yields impressive alpha matte prediction.

Motivated by this, we define two different feature flows in our framework (Figure \ref{fig:arch}): alpha feature flow (blue arrows) and image feature flow (yellow arrows). Alpha features are generated from the 6-channel input which is a concatenation of original image and trimap. The final alpha matte can be predicted directly from alpha features. Low-level image features contrast with the high-level alpha features. These features are generated only from the input image by a sequence of three convolutional layer with stride 2, which are analogous to the local color statistics in conventional affinity-based methods.

In other words, the alpha feature contains opacity information and low-level image feature contains appearance information. 
Given both opacity and appearance information, we can build an affinity graph and carry out opacity propagation as affinity-based methods.
Specifically, we utilize the low-level image feature to guide the information flow on alpha features.

\subsection{Guided Contextual Attention}

Inspired by the contextual attention for image inpainting proposed in \cite{yu2018generative}, we introduce our guided contextual attention block. 

As shown in Figure \ref{fig:gca}, the guided contextual attention leverages both the image feature and alpha feature. Firstly, the image feature are divided into known part and unknown part and $ 3\times3 $ patches are extracted from the whole image feature. Each feature patch represents the appearance information at a specific position. We reshape the patches as convolutional kernels. In order to measure the correlation between an unknown region patch $ U_{x,y} $ centered on $ (x,y)$ and an image feature patch $ I_{x', y'} $ centered on $ (x',y') $ , the similarity is defined as the normalized inner product:
\begin{equation}
	s_{(x,y), (x',y')} = 	
	\begin{cases}
	\lambda \quad &(x,y)=(x',y');\\
	\langle \frac{U_{x, y}}{\|U_{x, y}\|},\frac{I_{x', y'}}{\|I_{x', y'}\|}\rangle \quad &\mathrm{otherwise},	
	\end{cases}
\end{equation}
where $ U_{x, y}\in\mathcal{U}$ is also an element of the image feature patch set $ \mathcal{I} $, i.e. $ \mathcal{U} \subseteq \mathcal{I} $. The constant $ \lambda $ is a punishment hyperparameter that we use $ -10^4 $ in our model, which can avoid a large correlation between each unknown patch and itself.
In implementation, this similarity is computed by a convolution between unknown region features and kernels reshaped from image feature patches. Given the correlation, we carry out a scaled softmax along $ (x',y') $
dimension to attain the guided attention score for each patch as following, 
\begin{equation}
	a_{(x,y), (x',y')} = \mathrm{softmax}(w(\mathcal{U}, \mathcal{K}, x',y')s_{(x,y), (x',y')}),
\end{equation}
\begin{equation}
w(\mathcal{U}, \mathcal{K}, x',y') = \begin{cases}
\mathrm{clamp}(\sqrt{\frac{|\mathcal{U}|}{|\mathcal{K}|}}) \quad I_{x',y'} \in \mathcal{U};\\
\mathrm{clamp}(\sqrt{\frac{|\mathcal{K}|}{|\mathcal{U}|}}) \quad I_{x',y'} \in \mathcal{K},
\end{cases}
\label{eq:weight}
\end{equation}
\begin{equation}
\mathrm{clamp}(\phi) = \mathrm{min}(\mathrm{max}(\phi, 0.1), 10),
\end{equation}
in which  $ w(\cdot) $ is a weight function and $ \mathcal{K} = \mathcal{I}-\mathcal{U}$ is the set of image feature patches from known region. As distinct from image inpainting task, the area of unknown region in a  trimap is not under control. In many input trimaps, there are overwhelming unknown region and scarcely any known pixel. Thus, typically it is not feasible that only propagate the opacity information from the known region to unknown part. In our guided contextual attention, we let the unknown part borrow features from both known patches and unknown ones. Different weights are assigned to known and unknown patches based on the area of each region as the weight function defined in Eq. \eqref{eq:weight}. If the area of known region is larger, the known patches can convey more accurate appearance information which exposes the difference between foreground and background, hence we weigh known patches with a larger weight. Whereas,  if the unknown region has an overwhelming area, the known patches only provide some local appearance information, which may harm the opacity propagation. Then a small weight is assigned to known patches.

When we get guided attention scores from image features, we do the propagation on alpha features based on the affinity graph defined by guided attention.
Analogous to image features, patches are extracted and reshaped as filter kernels from alpha features. The information propagation is implemented as a deconvolution between guided attention scores and reshaped alpha feature patches. This deconvolution yields a reconstruction of alpha features in the unknown area and the values of overlapped pixels in the deconvolution are averaged. Finally, we combine the input alpha features and the propagation result by an element-wise summation. This element-wise summation works as a residual connection which can stable the training.

\begin{table}[t]
	\caption{The quantitative results on Composition-1k testing set. Best results are emphasized in bold. (- indicates not given in the original paper.)}
	\small
	\centering
	\begin{tabular}{lcccc}  
		\toprule
		Methods & MSE & SAD & Grad &Conn\\
		\midrule
		Learning Based Matting& 0.048&113.9 &91.6 &122.2\\
		Closed-Form Matting &0.091& 168.1 &126.9& 167.9\\
		KNN Matting& 0.103& 175.4& 124.1 &176.4\\     
		Deep Matting & 0.014&50.4& 31.0& 50.8\\
		IndexNet Matting &	0.013&45.8&	25.9&	43.7\\	
		SampleNet Matting&	0.0099 &40.35&	-&	-\\	
		\midrule
		Baseline   & 0.0106& 40.62 & 21.53 & 38.43\\
		Ours   &\textbf{0.0091}& \textbf{35.28}& \textbf{16.92} & \textbf{32.53}\\
		\bottomrule
	\end{tabular}
	\label{tab:adobe}
\end{table}

\begin{figure*}[t]
	\centering
	\includegraphics[width = 0.95\textwidth]{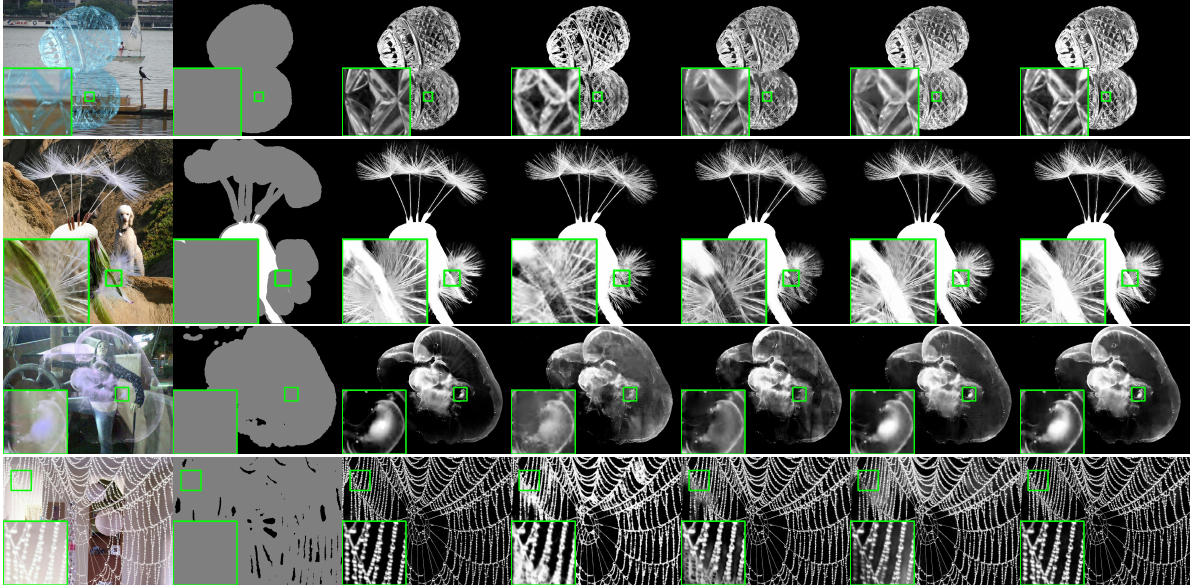}
	\caption{The visual comparison results on Adobe Composition-1k. From left to right, the original image, trimap, ground-truth, Deep Matting \cite{xu2017deep}, IndexNet Matting \cite{lu2019indices}, baseline and ours. }
	\label{fig:adobe}
\end{figure*}

\subsection{Network with Guided Contextual Attention}

Most of the affinity-based matting methods result in a closed-form solution based on the graph Laplacian \cite{levin2008closed,lee2011nonlocal,chen2013knn}. The closed-form solution can be seen as a fixed point of the propagation or a limitation of infinite propagation iterations \cite{zhou2004learning}.
Motivated by this, we stick two  guided contextual attention blocks to the encoder and decoder symmetrically in our stem. It aims to propagate more times in our model and take full advantage of the opacity information flow. 

When we compute the guided contextual attention on higher-resolution features, more detailed appearance information will be attended. However, on the other hand, the computational complexity of the attention block is $ O(c(hw)^2) $, where $ c, h, w $ are the channels, height and width of the feature map respectively. Therefore, we append two guided contextual attention blocks to the stage with $ 64\times64 $ feature maps.

The network of our GCA Matting is trained for $ 200,000 $ iterations with a batch size of 40 in total on the Adobe Image Matting dataset \cite{xu2017deep}. We perform optimization using Adam optimizer \cite{kingma2014adam} with $ \beta_1=0.5 $ and $ \beta_2=0.999 $. The learning rate is initialized to $ 4\times 10^{-4} $. Warmup and cosine decay \cite{loshchilov2016sgdr,goyal2017accurate,he2019bag} are applied to the learning rate. 


\section{Results}
In this section we report the evaluation results of our proposed model on two datasets, the Composition-1k testing set and alphamatting.com dataset. Both quantitative and qualitative results are shown in this section. We evaluate the quantitative results under the Sum of Absolute Differences (SAD), Mean Squared Error (MSE), Gradient error (Grad) and Connectivity error (Conn) proposed by \cite{rhemann2009perceptually}.

\subsection{Composition-1k Testing Dataset}

\begin{table*}[t]
	\caption{Our scores in the alpha matting benchmark, S, L and U denote the three trimap types, small, large and user, included in the benchmark. (Bold numbers indicate scores which rank the 1st place in the benchmark at the time of submission)}
	\scriptsize
	\centering
	\begin{tabular}{@{\;}l|@{}C{30pt}@{}|*{2}{@{}C{18pt}}@{}C{18pt}@{}*{8}{|@{}C{14pt}@{}C{14pt}@{}C{14pt}@{}}}  
		\toprule
		\multirow{2}{*}{Gradient Error} & \multicolumn{4}{c|}{Average Rank}&\multicolumn{3}{c|}{Troll}&\multicolumn{3}{c|}{Doll}&\multicolumn{3}{c|}{Donkey}&\multicolumn{3}{c|}{Elephant}&\multicolumn{3}{c|}{Plant}&\multicolumn{3}{c|}{Pineapple}&\multicolumn{3}{c|}{Plastic bag}&\multicolumn{3}{c}{Net}\\
		& Overall&S&L&U&S&L&U&S&L&U&S&L&U&S&L&U&S&L&U&S&L&U&S&L&U&S&L&U\\
		\midrule
		Ours&	\textbf{5.2}&	5&	4&	6.5&	0.1 &	\textbf{0.1} &	0.2 &	0.1 &	0.1 &	0.3 &	0.2 &	0.2 &	0.2 &	0.2 &	0.2 &	\textbf{0.3} &	1.3 &	1.6 &	1.9 &	0.7 &	0.8 &	1.4 &	\textbf{0.6} &	\textbf{0.7} &	\textbf{0.6} &	0.4 &	0.4 &	0.4 \\
		SampleNet Matting &	7.2&	3.6	&4.4	&13.6&\textbf{0.1} &	0.1 &	0.2 &	0.1 &	0.1 &	0.2 &	0.2 &	0.3 &	0.3 &	0.1 &	0.2 &	0.5 &	1.1 &	1.5 &	2.7 &	0.6 &	0.9 &	1 &	0.8 &	0.9 &	0.9 &	0.4 &	0.4 &	0.4\\
		IndexNet Matting&10.3&	8.6&	8.8&	13.6&	0.2 &	0.2 &	0.2 &	0.1 &	0.1 &	0.3 &	0.2 &	0.2 &	\textbf{0.2} &	0.2 &	0.2 &	0.4 &	1.7 &	1.9 &	2.5 &	1 &	1.1 &	1.3 &	1.1 &	1.2 &	1.2 &	0.4 &	0.5 &	0.5 \\
		AlphaGAN&14.9&	13.6&	12.5&	18.5&	0.2 &	0.2 &	0.2 &	0.2 &	0.2 &	0.3 &	0.2 &	0.3 &	0.3 &	0.2 &	0.2 &	0.4 &	1.8 &	2.4 &	2.7 &	1.1 &	1.4 &	1.5 &	0.9 &	1.1 &	1 &	0.5 &	0.5 &	0.6\\
		Deep Matting&15.6&	12&	12.3&	22.5&	0.4 &	0.4 &	0.5 &	0.2 &	0.2 &	\textbf{0.2} &	\textbf{0.1} &	\textbf{0.1} &	0.2 &	0.2 &	0.2 &	0.6 &	1.3 &	1.5 &	2.4 &	0.8 &	0.9 &	1.3 &	0.7 &	0.8 &	1.1 &	0.4 &	0.5 &	0.5 \\
		Information-flow matting&	18.3&	21.5&	16.5&	16.8&	0.2 &	0.2 &	0.2 &	0.2 &	0.2 &	0.4 &	0.4 &	0.4 &	0.4 &	0.3 &	0.4 &	0.4 &	1.7 &	1.8 &	2.2 &	0.9 &	1.3 &	1.3 &	1.5 &	1.4 &	0.8 &	0.5 &	0.6 &	0.5 \\
		\bottomrule
		\multicolumn{29}{c}{}\\
	\end{tabular}
	\label{tab:alphamatting}
\end{table*}

\begin{figure*}[t]
	\centering
	\includegraphics[width = \textwidth]{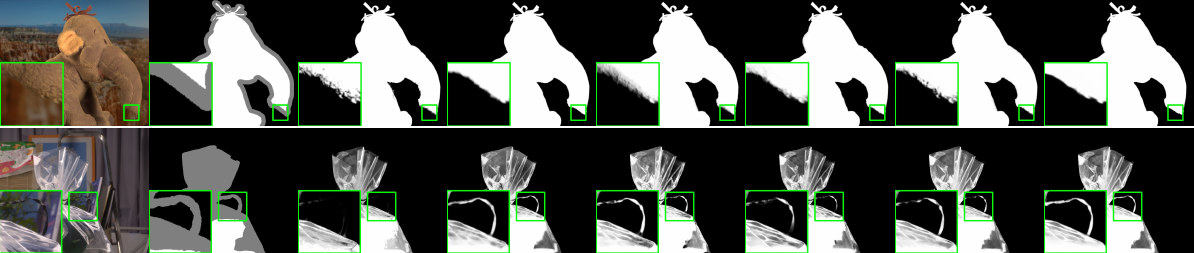}
	\caption{The alpha matte predictions of the test images from alphamatting.com benchmark. From left to right, the original image, trimap, Information-flow Matting \cite{aksoy2017designing} , Deep Matting \cite{xu2017deep}, AlphaGAN \cite{lutz2018alphagan}, IndexNet Matting \cite{lu2019indices}, SampleNet Matting	\cite{samplenet} and ours. }
	\label{fig:alphamatting}
\end{figure*}

The Composition-1k testing dataset proposed in \cite{xu2017deep} contains 1000 testing images which are composed from 50 foreground objects and 1000 different background images from Pascal VOC dataset \cite{everingham2015pascal}.

We compare our approach and the baseline model with three state-of-the-art deep image matting methods: Deep Matting \cite{xu2017deep}, IndexNet Matting \cite{lu2019indices} and SampleNet Matting \cite{samplenet}, as well as three conventional hand-crafted algorithms: Learning Based Matting \cite{zheng2009learning}, Closed-Form Matting \cite{levin2008closed} and KNN Matting \cite{chen2013knn}. The quantitative results are shown in Table \ref{tab:adobe}. Our method outperforms all of the state-of-the-art approaches. In addition, our baseline model also get better results than some of the top performing methods. The effectiveness of the proposed guided contextual attention  can be validated by the results displayed in Table \ref{tab:adobe}.

Some qualitative results are given in Figure \ref{fig:adobe}. The results of Deep Matting and IndexNet Matting are generated by source codes and pretrained models provided in \cite{lu2019indices}. As displayed in Figure \ref{fig:adobe}, our approach achieves better performance on different foreground objects, especially in the semitransparent regions. Advantages are more obvious with a larger unknown region.
This good performance  profits from the information flow between feature patches with similar appearance features.

Additionally, our proposed method can evaluate each image in Composition-1k testing dataset as a whole on a single Nvidia GTX 1080 with 8GB memory. Since we take each image as a whole in our network without scaling, the guided contextual attention blocks are applied to feature maps with a much higher resolution than $ 64\times 64 $ in training phase. This results in a better performance in the detailed texture.

\subsection{Alphamatting.com Benchmark dataset}

The alphamatting.com benchmark dataset \cite{rhemann2009perceptually} has eight different images. For each testing image, there are three corresponding trimaps, namely, "small", "large" and "user". The methods on the benchmark are ranked by the averaged rank over 24 alpha matte estimations in terms of four different metrics. We evaluate our method on the the alphamatting.com benchmark, and show the scores in Table \ref{tab:alphamatting}. Some top approaches in the benchmark are also displayed for comparison.

 As displayed in Table \ref{tab:alphamatting}, GCA Matting ranks the first place under the Gradient Error metric in the benchmark. The evaluation results of our method under the "large" and "user" trimaps are much better than the other top approaches.  The image matting becomes more difficult as the trimap has a larger unknown region. Therefore, we can say that our approach is more robust to changes in the area of unknown region. Additionally, our approach has almost the same overall ranks with the SampleNet under the MSE metric. Generally, the proposed GCA Matting is one of the top performing method on this benchmark dataset.

We provide some of the visual examples in Figure \ref{fig:alphamatting}. The results of our method and some top algorithms on "Elephant" and "Plastic bag" are displayed to demonstrate the good performance of our approach. For example, in the test image "Plastic bag", most of the previous methods make a mistake at the iron wire. However, our method learns from the contextual information in the surrounding background patches and predicts these pixels correctly.

\subsection{Visualization of Attention Map}

We visualize the attention map learned in the guided contextual attention block by demonstrating the pixel position with the largest attention score. Unlike the offset map widely used in optical flow estimation \cite{FlowNet,LiteFlowNet,PWC-Net} and image inpainting \cite{yu2018generative} which indicates the relative displacement of each pixel, our attention map demonstrates the absolute position of the corresponding pixel with highest attention activation. From this attention map, we can easily identify where the opacity information is propagated from for each feature pixel. As we can see in Figure \ref{fig:offset}, 
there is no information flow in the known region and 
feature patches in the unknown region tend to borrow information from the patches with similar appearance. 
Figure \ref{fig:offset} reveals where our GCA blocks attend to physically in the input image.
Since there is an adaption convolutional layer in the guided contextual attention block before patch extraction on image features, attention maps from two attention blocks are not identical. The weights of known and unknown part are shown in the top-left corner of the attention map.

From the attention offset map in Figure \ref{fig:offset}, we can easily recognize the car in the sieve. The light pink patches at the center of the sieve indicate that these features are propagated from the left part of the car. While blue patches show the features which are borrowed from the right-hand side road. These propagated features will assist in the identification of foreground and background in ensuing convolutional layers.

\section{Conclusions}
In this paper, we propose to solve the image matting problem by opacity information propagation in an end-to-end neural network.  Consequently, a guided contextual attention module is introduced to imitate the affinity-based propagation method by a fully convolutional manner. In the proposed attention module, the opacity information is transmitted  between alpha features under the guidance of appearance information. The evaluation results on both Composition-1k testing dataset and alphamatting.com dataset show the superiority of our proposed method.

\section{Acknowledgement}
This paper is supported by NSFC (No.61772330, 61533012, 61876109), the advanced research project (No.61403120201), Shanghai authentication key Lab. (2017XCWZK01), Technology Committee the interdisciplinary Program of Shanghai Jiao Tong University
(YG2015MS43). We also would like to thank the help and support from Versa.

{\small
\bibliographystyle{aaai}
\bibliography{10261.main}}

\begin{thebibliography}{}

\bibitem[\protect\citeauthoryear{Aksoy, Ozan~Aydin, and
  Pollefeys}{2017}]{aksoy2017designing}
Aksoy, Y.; Ozan~Aydin, T.; and Pollefeys, M.
\newblock 2017.
\newblock Designing effective inter-pixel information flow for natural image
  matting.
\newblock In {\em CVPR}.

\bibitem[\protect\citeauthoryear{Brock, Donahue, and
  Simonyan}{2019}]{brock2018large}
Brock, A.; Donahue, J.; and Simonyan, K.
\newblock 2019.
\newblock Large scale gan training for high fidelity natural image synthesis.
\newblock In {\em ICLR}.

\bibitem[\protect\citeauthoryear{Chen, Li, and Tang}{2013}]{chen2013knn}
Chen, Q.; Li, D.; and Tang, C.-K.
\newblock 2013.
\newblock Knn matting.
\newblock {\em IEEE TPAMI}.

\bibitem[\protect\citeauthoryear{Cho, Tai, and Kweon}{2016}]{cho2016natural}
Cho, D.; Tai, Y.-W.; and Kweon, I.
\newblock 2016.
\newblock Natural image matting using deep convolutional neural networks.
\newblock In {\em ECCV}.

\bibitem[\protect\citeauthoryear{Dosovitskiy \bgroup et al\mbox.\egroup
  }{2015}]{FlowNet}
Dosovitskiy, A.; Fischer, P.; Ilg, E.; H{\"{a}}usser, P.; Hazirbas, C.; Golkov,
  V.; van~der Smagt, P.; Cremers, D.; and Brox, T.
\newblock 2015.
\newblock Flownet: Learning optical flow with convolutional networks.
\newblock In {\em ICCV}.

\bibitem[\protect\citeauthoryear{Everingham \bgroup et al\mbox.\egroup
  }{2015}]{everingham2015pascal}
Everingham, M.; Eslami, S.~A.; Van~Gool, L.; Williams, C.~K.; Winn, J.; and
  Zisserman, A.
\newblock 2015.
\newblock The pascal visual object classes challenge: A retrospective.
\newblock {\em International journal of computer vision} 111(1):98--136.

\bibitem[\protect\citeauthoryear{Feng, Liang, and
  Zhang}{2016}]{feng2016cluster}
Feng, X.; Liang, X.; and Zhang, Z.
\newblock 2016.
\newblock A cluster sampling method for image matting via sparse coding.
\newblock In {\em European Conference on Computer Vision},  204--219.
\newblock Springer.

\bibitem[\protect\citeauthoryear{Gastal and Oliveira}{2010}]{gastal2010shared}
Gastal, E.~S., and Oliveira, M.~M.
\newblock 2010.
\newblock Shared sampling for real-time alpha matting.
\newblock In {\em Computer Graphics Forum}, volume~29,  575--584.
\newblock Wiley Online Library.

\bibitem[\protect\citeauthoryear{Goyal \bgroup et al\mbox.\egroup
  }{2017}]{goyal2017accurate}
Goyal, P.; Doll{\'a}r, P.; Girshick, R.; Noordhuis, P.; Wesolowski, L.; Kyrola,
  A.; Tulloch, A.; Jia, Y.; and He, K.
\newblock 2017.
\newblock Accurate, large minibatch sgd: Training imagenet in 1 hour.
\newblock {\em arXiv preprint arXiv:1706.02677}.

\bibitem[\protect\citeauthoryear{He \bgroup et al\mbox.\egroup
  }{2011}]{he2011global}
He, K.; Rhemann, C.; Rother, C.; Tang, X.; and Sun, J.
\newblock 2011.
\newblock A global sampling method for alpha matting.
\newblock In {\em CVPR 2011},  2049--2056.
\newblock IEEE.

\bibitem[\protect\citeauthoryear{He \bgroup et al\mbox.\egroup
  }{2016}]{he2016deep}
He, K.; Zhang, X.; Ren, S.; and Sun, J.
\newblock 2016.
\newblock Deep residual learning for image recognition.
\newblock In {\em CVPR}.

\bibitem[\protect\citeauthoryear{He \bgroup et al\mbox.\egroup
  }{2019}]{he2019bag}
He, T.; Zhang, Z.; Zhang, H.; Zhang, Z.; Xie, J.; and Li, M.
\newblock 2019.
\newblock Bag of tricks for image classification with convolutional neural
  networks.
\newblock In {\em CVPR}.

\bibitem[\protect\citeauthoryear{Hui, Tang, and Loy}{2018}]{LiteFlowNet}
Hui, T.; Tang, X.; and Loy, C.~C.
\newblock 2018.
\newblock Liteflownet: {A} lightweight convolutional neural network for optical
  flow estimation.
\newblock In {\em CVPR},  8981--8989.

\bibitem[\protect\citeauthoryear{Ioffe and Szegedy}{2015}]{ioffe2015batch}
Ioffe, S., and Szegedy, C.
\newblock 2015.
\newblock Batch normalization: Accelerating deep network training by reducing
  internal covariate shift.
\newblock {\em arXiv preprint arXiv:1502.03167}.

\bibitem[\protect\citeauthoryear{Isola \bgroup et al\mbox.\egroup
  }{2017}]{isola2017image}
Isola, P.; Zhu, J.-Y.; Zhou, T.; and Efros, A.~A.
\newblock 2017.
\newblock Image-to-image translation with conditional adversarial networks.
\newblock In {\em CVPR}.

\bibitem[\protect\citeauthoryear{Kingma and Ba}{2014}]{kingma2014adam}
Kingma, D.~P., and Ba, J.
\newblock 2014.
\newblock Adam: A method for stochastic optimization.
\newblock {\em arXiv preprint arXiv:1412.6980}.

\bibitem[\protect\citeauthoryear{Kipf and Welling}{2016}]{kipf2016semi}
Kipf, T.~N., and Welling, M.
\newblock 2016.
\newblock Semi-supervised classification with graph convolutional networks.
\newblock {\em arXiv preprint arXiv:1609.02907}.

\bibitem[\protect\citeauthoryear{Lee and Wu}{2011}]{lee2011nonlocal}
Lee, P., and Wu, Y.
\newblock 2011.
\newblock Nonlocal matting.
\newblock In {\em CVPR 2011},  2193--2200.
\newblock IEEE.

\bibitem[\protect\citeauthoryear{Levin, Lischinski, and
  Weiss}{2008}]{levin2008closed}
Levin, A.; Lischinski, D.; and Weiss, Y.
\newblock 2008.
\newblock A closed-form solution to natural image matting.
\newblock {\em IEEE TPAMI}.

\bibitem[\protect\citeauthoryear{Li \bgroup et al\mbox.\egroup
  }{2019}]{li2019inductive}
Li, Y.; Zhang, J.; Zhao, W.; and Lu, H.
\newblock 2019.
\newblock Inductive guided filter: Real-time deep image matting with weakly
  annotated masks on mobile devices.
\newblock {\em arXiv preprint arXiv:1905.06747}.

\bibitem[\protect\citeauthoryear{Lin \bgroup et al\mbox.\egroup
  }{2014}]{lin2014microsoft}
Lin, T.-Y.; Maire, M.; Belongie, S.; Hays, J.; Perona, P.; Ramanan, D.;
  Doll{\'a}r, P.; and Zitnick, C.~L.
\newblock 2014.
\newblock Microsoft coco: Common objects in context.
\newblock In {\em ECCV},  740--755.
\newblock Springer.

\bibitem[\protect\citeauthoryear{Liu \bgroup et al\mbox.\egroup
  }{2018}]{liu2018image}
Liu, G.; Reda, F.~A.; Shih, K.~J.; Wang, T.-C.; Tao, A.; and Catanzaro, B.
\newblock 2018.
\newblock Image inpainting for irregular holes using partial convolutions.
\newblock In {\em Proceedings of the European Conference on Computer Vision
  (ECCV)},  85--100.

\bibitem[\protect\citeauthoryear{Long, Shelhamer, and
  Darrell}{2015}]{long2015fully}
Long, J.; Shelhamer, E.; and Darrell, T.
\newblock 2015.
\newblock Fully convolutional networks for semantic segmentation.
\newblock In {\em Proceedings of the IEEE conference on computer vision and
  pattern recognition},  3431--3440.

\bibitem[\protect\citeauthoryear{Loshchilov and
  Hutter}{2016}]{loshchilov2016sgdr}
Loshchilov, I., and Hutter, F.
\newblock 2016.
\newblock Sgdr: Stochastic gradient descent with warm restarts.
\newblock {\em arXiv preprint arXiv:1608.03983}.

\bibitem[\protect\citeauthoryear{Lu \bgroup et al\mbox.\egroup
  }{2019}]{lu2019indices}
Lu, H.; Dai, Y.; Shen, C.; and Xu, S.
\newblock 2019.
\newblock Indices matter: Learning to index for deep image matting.
\newblock In {\em ICCV}.

\bibitem[\protect\citeauthoryear{Lutz, Amplianitis, and
  Smolic}{2018}]{lutz2018alphagan}
Lutz, S.; Amplianitis, K.; and Smolic, A.
\newblock 2018.
\newblock Alphagan: Generative adversarial networks for natural image matting.
\newblock In {\em BMVC}.

\bibitem[\protect\citeauthoryear{Miyato \bgroup et al\mbox.\egroup
  }{2018}]{miyato2018spectral}
Miyato, T.; Kataoka, T.; Koyama, M.; and Yoshida, Y.
\newblock 2018.
\newblock Spectral normalization for generative adversarial networks.
\newblock {\em arXiv preprint arXiv:1802.05957}.

\bibitem[\protect\citeauthoryear{Rhemann \bgroup et al\mbox.\egroup
  }{2009}]{rhemann2009perceptually}
Rhemann, C.; Rother, C.; Wang, J.; Gelautz, M.; Kohli, P.; and Rott, P.
\newblock 2009.
\newblock A perceptually motivated online benchmark for image matting.
\newblock In {\em CVPR}.

\bibitem[\protect\citeauthoryear{Ronneberger, Fischer, and
  Brox}{2015}]{ronneberger2015u}
Ronneberger, O.; Fischer, P.; and Brox, T.
\newblock 2015.
\newblock U-net: Convolutional networks for biomedical image segmentation.
\newblock In {\em MICCAI}.

\bibitem[\protect\citeauthoryear{Sun \bgroup et al\mbox.\egroup
  }{2018}]{PWC-Net}
Sun, D.; Yang, X.; Liu, M.; and Kautz, J.
\newblock 2018.
\newblock Pwc-net: Cnns for optical flow using pyramid, warping, and cost
  volume.
\newblock In {\em CVPR}.

\bibitem[\protect\citeauthoryear{Tang \bgroup et al\mbox.\egroup
  }{2019}]{samplenet}
Tang, J.; Aksoy, Y.; \"Oztireli, C.; Gross, M.; and Ayd{\i}n, T.~O.
\newblock 2019.
\newblock Learning-based sampling for natural image matting.
\newblock In {\em Proc. CVPR}.

\bibitem[\protect\citeauthoryear{Vaswani \bgroup et al\mbox.\egroup
  }{2017}]{vaswani2017attention}
Vaswani, A.; Shazeer, N.; Parmar, N.; Uszkoreit, J.; Jones, L.; Gomez, A.~N.;
  Kaiser, {\L}.; and Polosukhin, I.
\newblock 2017.
\newblock Attention is all you need.
\newblock In {\em NIPS}.

\bibitem[\protect\citeauthoryear{Veli{\v{c}}kovi{\'c} \bgroup et
  al\mbox.\egroup }{2017}]{velivckovic2017graph}
Veli{\v{c}}kovi{\'c}, P.; Cucurull, G.; Casanova, A.; Romero, A.; Lio, P.; and
  Bengio, Y.
\newblock 2017.
\newblock Graph attention networks.
\newblock {\em arXiv preprint arXiv:1710.10903}.

\bibitem[\protect\citeauthoryear{Wang and Cohen}{2007}]{wang2007optimized}
Wang, J., and Cohen, M.~F.
\newblock 2007.
\newblock Optimized color sampling for robust matting.
\newblock In {\em 2007 IEEE Conference on Computer Vision and Pattern
  Recognition},  1--8.
\newblock IEEE.

\bibitem[\protect\citeauthoryear{Wang \bgroup et al\mbox.\egroup
  }{2018}]{wang2018non}
Wang, X.; Girshick, R.; Gupta, A.; and He, K.
\newblock 2018.
\newblock Non-local neural networks.
\newblock In {\em Proceedings of the IEEE Conference on Computer Vision and
  Pattern Recognition},  7794--7803.

\bibitem[\protect\citeauthoryear{Wang, Cohen, and others}{2008}]{wang2008image}
Wang, J.; Cohen, M.~F.; et~al.
\newblock 2008.
\newblock Image and video matting: a survey.
\newblock {\em Foundations and Trends{\textregistered} in Computer Graphics and
  Vision} 3(2):97--175.

\bibitem[\protect\citeauthoryear{Xu \bgroup et al\mbox.\egroup
  }{2017}]{xu2017deep}
Xu, N.; Price, B.; Cohen, S.; and Huang, T.
\newblock 2017.
\newblock Deep image matting.
\newblock In {\em CVPR}.

\bibitem[\protect\citeauthoryear{Yang \bgroup et al\mbox.\egroup
  }{2019}]{yang2019xlnet}
Yang, Z.; Dai, Z.; Yang, Y.; Carbonell, J.; Salakhutdinov, R.; and Le, Q.~V.
\newblock 2019.
\newblock Xlnet: Generalized autoregressive pretraining for language
  understanding.
\newblock {\em arXiv preprint arXiv:1906.08237}.

\bibitem[\protect\citeauthoryear{Yu \bgroup et al\mbox.\egroup
  }{2018}]{yu2018generative}
Yu, J.; Lin, Z.; Yang, J.; Shen, X.; Lu, X.; and Huang, T.~S.
\newblock 2018.
\newblock Generative image inpainting with contextual attention.
\newblock In {\em CVPR}.

\bibitem[\protect\citeauthoryear{Zhang \bgroup et al\mbox.\egroup
  }{2019}]{zhang2018self}
Zhang, H.; Goodfellow, I.; Metaxas, D.; and Odena, A.
\newblock 2019.
\newblock Self-attention generative adversarial networks.
\newblock In {\em ICML}.

\bibitem[\protect\citeauthoryear{Zheng and
  Kambhamettu}{2009}]{zheng2009learning}
Zheng, Y., and Kambhamettu, C.
\newblock 2009.
\newblock Learning based digital matting.
\newblock In {\em ICCV}.

\bibitem[\protect\citeauthoryear{Zhou \bgroup et al\mbox.\egroup
  }{2004}]{zhou2004learning}
Zhou, D.; Bousquet, O.; Lal, T.~N.; Weston, J.; and Sch{\"o}lkopf, B.
\newblock 2004.
\newblock Learning with local and global consistency.
\newblock In {\em Advances in neural information processing systems},
  321--328.

\end{thebibliography}
\end{document}